\documentclass{article}

\usepackage{arxiv}
\usepackage{scrextend}
\KOMAoptions{fontsize=12pt}

\usepackage[utf8]{inputenc} % allow utf-8 input
\usepackage[T1]{fontenc}    % use 8-bit T1 fonts
\usepackage{hyperref}       % hyperlinks
\hypersetup{colorlinks=true, allcolors=blue}
\usepackage{url}            % simple URL typesetting
\usepackage{booktabs}       % professional-quality tables
\usepackage{amsfonts}       % blackboard math symbols
\usepackage{nicefrac}       % compact symbols for 1/2, etc.
\usepackage{microtype}      % microtypography
\usepackage{lipsum}		% Can be removed after putting your text content
\usepackage{graphicx}
\usepackage[round]{natbib}
\usepackage{doi}
\usepackage{lmodern}
\usepackage[ruled,vlined,english]{algorithm2e}
\usepackage[ruled,vlined]{algorithm2e}

\usepackage{adjustbox}
\usepackage{threeparttable}

\SetCommentSty{mycommfont}
\usepackage{amsmath}

\usepackage{float}
\usepackage[labelfont=bf, justification=justified]{caption}
\usepackage{subcaption} % for 'subfigure' environment

\title{1D Convolutional neural networks and machine learning algorithms for spectral data classification with a case study for Covid-19}

\date{} 					% Or removing it

\author{Breno Aguiar Krohling \\
	Nature-inspired Computing Lab \\
	LABCIN - UFES \\
	Federal University of Espirito Santo \\
	Vitória, Brazil \\
	\texttt{brenokrohling@gmail.com} \\
	\And
	Renato A. Krohling \\
	Nature-inspired Computing Lab \\
	LABCIN - UFES \\
	PPGI - UFES \\
	Federal University of Espirito Santo \\
	Vitória, Brazil \\
	\texttt{krohling.renato@gmail.com} \\
}

\begin{document}
\maketitle

\begin{abstract}
	Machine and deep learning algorithms have increasingly been applied to solve problems in various areas of knowledge. Among these areas, Chemometrics has been benefited from the application of these algorithms in spectral data analysis. Commonly, algorithms such as Support Vector Machines and Partial Least Squares are applied to spectral datasets to perform classification and regression tasks. In this paper, we present a 1D convolutional neural networks (1D-CNN) to evaluate the effectiveness on spectral data obtained from spectroscopy. In most cases, the spectrum signals are noisy and present overlap among classes.  Firstly, we perform  extensive experiments including 1D-CNN compared to machine learning algorithms and standard algorithms used in Chemometrics on spectral data classification for the most known datasets available in the literature. Next, spectral samples of the SARS-COV2 virus, which causes the COVID-19, have recently been collected via spectroscopy  was used as a case study. Experimental results indicate superior performance of 1D-CNN over machine learning algorithms and standard algorithms, obtaining an average accuracy of $96.5\%$, specificity of $98\%$, and sensitivity of $94\%$. The promissing  obtained results indicate the feasibility to use 1D-CNN in automated systems to diagnose COVID-19 and other viral diseases in the future.
\end{abstract}

% keywords can be removed

%%% Introduction
\section{Introduction}

The use of spectroscopy has been showing an increasing research interest and practical applications with the advancement of spectral measurement techniques and data analysis studies in the field of chemometrics. In particular, the use of measurements in the infrared (IR) and near and infrared (NIR) spectrum, which provides information that is not visible by simple observation. The use of infrared spectroscopy is considered one of the most important tools in chemometrics \citep{stuart2000infrared}, as it has a low cost, is fast, and can analyze samples in their most diverse states, making it also a non-invasive approach. Besides that, the use of machine learning and deep learning, such as artificial neural networks, have been introduced in areas previously dominated by statistical analysis methods and/or classical artificial intelligence algorithms. With the growth of neural network applications in the last few years, such algorithms began to be explored in problems of different natures, including spectroscopy problems.

Data from spectroscopy presents some problems such as high dimensionality, and small datasets. Thus, It usually requires the use of pre processing techniques such as noise filtering and dimensionality reduction for proper classification. However, artificial neural networks (NN) are known to deal well with high dimensionality and have achieved remarkable results  for different applications. However, NN need a significant amount of data to work properly. In such cases, it becomes necessary to analyze the trade-off between using standard methods or migrate to new approaches.

Convolutional neural networks can be applied to challenging real-world problems, such as the diagnosis of the Sars-Cov-2 virus. With the advance of the COVID-19 pandemic, testing the population in large scale for sanitary and public health reasons became necessary. Sars-cov2 virus testing presents two main barriers: 1) the price of testing, and 2) time to get the results. While the first affects mostly underdeveloped countries, the second influences containment plans around the world. By testing with information acquired via spectrometers and processing the data through a robust artificial intelligence model, information about the patient's diagnosis could be obtained quickly and at a low cost, making it accessible to public and private health services.

 With the advances in machine learning techniques, there is a tendency to migrate from standard algorithms used in spectroscopy such as Support Vector Machines and Partial Least Squares, to recent approaches such as Convolutional Neural Networks \citep{zhang2019deepspectra} and \citep{yuanyuan:2018quantitative}. However, CNNs commonly require a large amount of data to train the model. Although there are public databases  \citep{kaewseekhao2020dataset}, \citep{chauvergne2020dataset} and \citep{zyubin2020dataset} there is still a lack of spectral datasets in public domain. So, it is necessary to investigate the performance of such algorithms when applied to problems with a small number of instances of training. Furthermore, spectral data present some characteristics, such as overlap of samples at certain wavelengths, which make the classification task difficult, as in the COVID-19 dataset provided by \cite{yin2021efficient}.

The main contribution of this work are twofold:
\begin{itemize}
\item we applied a 1D-CNN to 7 available public spectral datasets in order to validate the results as compared to conventional machine learning algorithms. 
\item  we present a case study involving spectral data of COVID-19 and show the feasibility of 1D-CNN.
\end{itemize}

This reminder of this paper is organized as follows. In section 2, we present related works. In section 3,  the methodology  involving pre-processing techniques and machine learning algorithms as well as a 1D convolutional neural network are described. The results are presented and   a comparative analysis is carried out with discusions in section 4, and in section 5 the article is concluded with directions for future research.

\section{Related Works}

Sepctroscopy techniques are used in several areas of knowledge and can be applied to approach several real-world problems, from the classification \textit{in vivo} of skin lesions \citep{mcintosh:2001} to determine the quality of agricultural products \citep{hayati2020enhanced}. \cite{mcintosh:2001} addressed the task of characterization skin lesions through statistical methods, specifically the use of the paired \textit{t-test}. a classification of carcinogenic lesions was performed using Linear Discriminant Analysis (LDA), showing high accuracy and feasibility for discriminating skin lesions in a non-invasive way. 

\cite{gniadecka:2004} combined spectroscopy techniques and Multilayer Perceptron neural networks for early diagnosis of skin cancer. The study aimed to differentiate melanoma, a skin cancer with a higher letal rate, from other skin lesions, such as pigmented nevus and basal cell carcinoma. The influence of chemical changes in cancer tissue and their importance for classification was investigated. The results obtained showed high sensitivity and specificity, important metrics in the detection of melanoma.
\cite{xiang:2010} used infrared spectral data to diagnose endometrial cancer. The study contained 77 samples, including carcinogenic material and a control group. Noise reduction techniques were used as the Savitsky-Golay filter and data dimensionality reduction with PCA. Post-processed data were used for training artificial neural networks and the obtained results showed effectiveness for early diagnosis of the disease.

\cite{wu:2012} used spectroscopy data to non-invasively determine the presence of proteins and polysaccharides in powder samples of \textit{coriolus versicolor}, a medicinal mushroom species. The authors compare artificial neural networks applied to original data and data with reduced dimensionality by applying PCA. The neural network was trained with the \textit{Backpropagation} algorithm and presented better results using data with reduced dimensionality.
\cite{liu:2017} used Deep Auto-Encoder (DAE) to extract and reduce features from the infrared spectral data. To evaluate its effectiveness, the algorithm was applied to the Cigarettes dataset and compared with methods established in the literature, such as PCA and in turn, the KNN  algorithm was applied. The results showed that the model using DAE was able to extract features with better differentiation capacity as the concurrent approaches.
%\cite{zhao:2017} propuseram um sistema \textit{real-time} para diagnóstico %de lesões de pele. Foi introduzido uma base de dados de espectroscopia contendo %518 amostras distribuídas em 10 classes. A curva ROC e sensitividade obtidas foram, respectivamente, $0.879$ e $0.90$.   
%esse paper é uma boa citacao para introducao de espectroscopia, bem detalhado historico e tecnico%%

\cite{peng:2018} proposed an algorithm for extracting spectral information in the frequency domain. Principal components were determined by the entropic contribution of each component together with genetic algorithm. The selected components were used to build a regression model with Partial Least Square (PLS), showing a performance improvement when compared to the results obtained with the original data.

\cite{cui:2018modern} performed tests on three spectroscopy datasets to demonstrate the effectiveness of CNNs in relation to the PLS Regression method. The results obtained showed that the convolution layer of a CNN can act as a  preprocessor of the spectrum, not requiring the application of signal processing techniques. In addition, results demonstrated the superiority of CNN in terms of the metrics used to validate the model, even using datasets with a low number of samples.

\cite{yuanyuan:2018quantitative} applied an ensemble of CNNs to the Corn, Gasoline, and Mixed Gases datasets and the approach was compared with classical methods in the spectroscopy field, such as PLS, as well as neural networks trained with Backpropagation. Results showed that the proposed model obtained superior performance in all case studies investigated, showing its relevance for spectrum classification.

\cite{lima:2019} investigated the discrimination between non-melanoma cancer (basal cell carcinoma and squamous cell carcinoma) and actinic keratosis and normal skin. The study used spectral data collected from in-vivo and ex-vivo tissues. The t-test was used to find regions of the spectrum with significant differences between samples. The selected regions were used to discriminate the lesions using Euclidean and Mahalanobis distance. Results obtained show robustness when using ex-vivo tissue data.

\cite{seoni:2019} investigated hemoglobin concentration using infrared spectrum data on normal skin tissue and skin tissue with actinic keratosis, a benign skin lesion that, without treatment, can evolve into a cancerous lesion. The MANOVA statistical analysis was applied to assess possible differences between the healthy skin group and the one containing actinic keratosis. The results showed that, in general, it was possible to differentiate the two groups through the statistical approach. 

\cite{zhang2019deepspectra} introduced a new approach based on a CNN with 1D convolutions and the adoption of Inception structure in two convolution layers. The method was evaluated in four public spectroscopy datasets: Corn, Tablet, Wheat, and Soil. The results obtained were superior to other CNN architectures and linear methods such as PLS and Support Vector Regression (SVR). In addition, pre-processing methods applied to the raw spectrum signal have also been applied in order to investigate the influence on the prediction results. The authors concluded that signal noise reduction is a necessary step to improve the performance of the algorithms.

\cite{chen:2019:end} evaluated the ability to extract features from a CNN without applying dimensionality reduction on the spectrum. Experiments were carried out using the dataset Corn and compared with traditional chemometric methods such as PLS and neural network training algorithms such as Extreme Learning Machine (ELM) and Backpropagation. The results obtained in the tests performed show that: 1) If the methods/algorithms use the original spectrum as input, the CNN obtains better performance; and 2) If standard methods/algorithms are combined with dimensionality reduction algorithms and variable selection, the results obtained are similar to those obtained by CNN, with no statistical difference between the methods.

\cite{chen:2020} proposed the use of ensemble of artificial neural networks trained with the ELM algorithm. The proposed algorithm uses random initialization of weights of the hidden layer to obtain diversity among models. The method was tested in 3 datasets, Tecator, Shootcut\_2002, and Tablet. The approach was compared with PLS, and was superior as compared to those achieved by classic chemometric algorithm.
\cite{yan:2020} collected infrared spectrum data in the hydrolysis process of materials used in traditional Chinese medicine. Regression models were developed using CNN and PLS with real-time data. The proposed CNN architecture presented better results over PLS, but with higher processing cost and computational time. Therefore, for real-time applications, it is necessary to analyze the maximum latency of the interval between results to decide which method should be used.

\section{Methodology}
%\subsection{Chemometrics}
\subsection{Data pre-processing }
%Chemometrics is a subfield of chemistry known to embrace analytical chemistry in conjunction with statistics, mathematical modeling and computer science \citep{gemperline2006practical}. Among the data collection/analysis methods used in chemometrics, spectroscopy has a relevant position due to its ease of use and accessibility.
%
Spectroscopy is a technique based on the vibrations of atoms in a molecule. The spectrum can be analyzed and classified by its wavelength, such as ultraviolet, visible, and infrared. Each wavelength provides unique information for analyzing a sample, so it is necessary to know the best wavelength range to use when collecting data. For example, the spectrum obtained at the infrared wavelength is widely used in applications that need to identify organic components of a sample \citep{stuart2000infrared}. The data generated by this technique, known as a spectrum, is obtained by emitting an amount of radiation onto a sample and determining which part of the radiation is absorbed or reflected by the sample at a given energy level \citep{stuart2000infrared}.

Data collected through spectrometers can suffer different types of interference, from low-quality equipment to external interference in uncontrolled environments, such as light scaterring, and noisy data, among others. Noisy components of the the spectrum signal is a challenging issue in spectral analysis, as the instability of the collected signal can lead to wrong conclusions. This instability can be reduced by performing more than one reading of the sample, obtaining the average spectra, but without eliminating all the existing noise. As a result, signal processing techniques become a crucial step in spectral analysis to remove, or reduce undesired noisy components of the spectral signal.

Selecting an appropriate technique to perform data pre-processing can directly influence the model performance and data interpretability \citep{lee2017contemporary}. The most used pre-processing techniques can be divided into two categories: 1) Scatter-correction, and 2) Spectral derivatives \citep{rinnan2009review}. Scatter-correction algorithms are used to reduce the variability between samples. Spectral derivates algorithms use convolution operations, usually by fitting  a polynomial to the points of the spectrum. 

In this work, we use one of the most standard pre-processing filter in spectral analysis, the Savitzky-Golay (SG) filter \citep{savitzky1964smoothing}, and evaluate its influence on the classification results. \cite{savitzky1964smoothing} proposed an algorithm to smooth data based on polynomial interpolation. They demonstrated that fitting a polynomial to a set of points in the spectrum and then evaluating it at the subsequent point respecting the point approximation interval would be equivalent to more complex convolution operations. Furthermore, they showed that the algorithm can remove noise from the data and keep the shape and peaks in the spectrum wavelengths, preserving original characteristics in the smoothed data. %A Figura \ref{fig-sg} ilustra um exemplo da aplicação do filtro de SG para eliminar o ruído de um sinal.

%\begin{figure}[t]
%  \begin{subfigure}{0.5\textwidth}
%    \includegraphics[width=\textwidth]{figuras/SG/TM_WO_SG.png}
%    \caption{Sinal ruidoso.}
%  \end{subfigure}
  %
%  \begin{subfigure}{0.5\textwidth}
%    \includegraphics[width=\textwidth]{figuras/SG/TM_W_SG.png}
%    \caption{Sinal ruidoso e suavizado pelo filtro de Savitzky-Golay.}
%  \end{subfigure}
%    \caption{Ilustrção (a) de um sinal ruidoso e (b) após a aplicação do filtro de Savitzky-Golay para suavizá-lo.}
%  \label{fig-sg}
%\end{figure}

\subsection{Machine Learning}
One of the fastest growing areas in computing is Artificial Intelligence. Its applications are present in medicine, engineering, finance, among other areas \citep{pannu2015artificial}. One of the sub-areas of Artificial Intelligence is machine learning, which consists of algorithms acquiring knowledge from already known data and, subsequently, being able to infer solutions for new instances of the problem without being explicitly programmed for it \citep{ alpaydin2020introduction}. %Machine learning problems can be grouped according to their characteristics, taking into account available data and its purpose.
Supervised Learning problems can be considered the most common among the existing ones. The algorithms look for patterns in order to learn how to associate an input $\textbf{x}$ to its respective output $y$. In general, supervised learning algorithms learn through examples how to classify an output $y$ given an input $\textbf{x}$ by estimating $p(y | \textbf{x})$ \citep{goodfellow2016deep}. The machine learning algorithms used in this work are Partial Least Square - Discriminant Analysis, Support Vector Machine, k-Nearest Neighbors \citep{alpaydin2020introduction}, and Many-objective clustering based on Hill Climbing and minimum spanning tree \citep{esgario2018clustering}.
%They are shortly described in Appendix A.

%Unlike what happens in supervised learning, we may face problems in which the dataset does not have the desired outputs, i.e., they are not labeled. Therefore, the algorithms try to group the data autonomously, according to observed characteristics. In this case, it handles of unsupervised learning.
%and unsupervised algorithms seek to estimate the probability distribution $p(\textbf{x})$ that generated the presented data 

%%% Methodology
\subsection{1D Convolutional Neural Networks}
Artificial neural networks (ANN) are structures composed of simple computing units, called artificial neurons, interconnected with each other \citep{schalkoff1997artificial}. The connections between neurons form complex layers for computing and propagating information. The algorithm for training ANN's is known as \textit{Backpropagation}. Through it, it is possible to train ANNs for classification and regression/prediction tasks. Standard ANNs are characterized by three main layers, the data input layer, the hidden layers, and the output layer.

Convolutional neural networks (CNN), initially proposed by \cite{cnn-1989}, are specialized networks for applications that present data in grid format such as images and temporal series, as they are able to capture spatial and temporal features \citep{goodfellow2016deep}. The architecture gained prominence in 2012 when a CNN-based model won the LSVRC-12 competition in the ImageNet database \citep{krizhevsky2012imagenet}. The main difference between CNN's and traditional ANN's is the presence of convolution and pooling layers in their architecture. CNNs use convolution operations instead of matrix operations in at least one of their layers \citep {goodfellow2016deep}. The main components of a CNN and its work principle during the training process are described in the following. %Como esse trabalho foca em séries temporais, será descrito a CNN em uma dimensão. %Para o leitor interessado em CNN para classificação de imagens refira-se a \cite{goodfellow2016deep}.

\paragraph{Convolution Layer:} Convolutions are linear operations on two functions, that generate a third one. In CNN's, convolutions are applied to extract features, where the first function of a convolution is represented by the input data of the network, and the second is represented by a \textit{kernel}, a sliding filter used to extract features. By applying the convolution operation between these two functions, it is generated feature maps, which contain relevant information from the original input data. CNN's can have layers of convolutions of one, two, or three dimensions, depending on the problem domain and data characteristics. For example, image recognition applications use 2D convolutions, while applications whose data comes from time-series use 1D convolution.

%Ao adicionarmos camadas de convolução na arquitetura de uma rede neural, obtemos algumas características importantes, interações esparsas e compartilhamento de parâmetros.
%Segundo \cite{goodfellow2016deep}, as interações esparsas se dão devido ao fato de reduzirmos as operações matriciais das unidades da rede ao utilizarmos camadas de convolução, reduzindo a quantidade de parâmetros gerados entre as camadas de entrada e saída e, consequentemente, impactando no desempenho computacional do modelo e eficiência estatística. O compartilhamento de parâmetros ocorre pois no processo de convolução um parâmetro é utilizado em mais de um função do modelo devido a aplicação deslizante do \textit{kernel}.

After convolution operations, feature maps are applied to an activation function. The activation function aims to add non-linearity to the model \citep{goodfellow2016deep}. When choosing an activation function, some characteristics must be observed, such as: The computational cost must be small, since the operation is performed frequently. Due to the training process of a neural network is based on gradient descent, the activation function must also be differentiable. The Rectified Linear Unit (ReLU) activation function is commonly used in neural networks because it presents such characteristics. The function returns zero for all negative values and the value itself for positive values. It is described by $ReLU(x) = max(0,x)$.

\paragraph{Pooling Layer:} The pooling layer, normally applied after the convolution operation, consists of reducing the dimensionality of the feature map in order to eliminate redundant information and keep the main features \citep{goodfellow2016deep}. Similar to convolution, a sliding window moves over the feature map generating a new output. However, pooling operations are deterministic, typically calculating the maximum or average value of the elements in the sliding window \citep{zhang2020dive}. When we apply the MaxPooling technique, the maximum value present in the analyzed data group is propagated to the next layers, while the AvgPooling technique propagates the average value of the data.

\paragraph{Fully connected layer:} The combination of convolution and pooling layers alone is not enough to perform the classification/regression task, its main function in the architecture is to extract features from the given data. The extracted features, therefore, need a classifier to complete the task, which we call a fully connected layer. Typically, the fully connected layer of a CNN is represented by an ANN known as a Multilayer Perceptron (MLP) \citep{zhang2020dive}, a feedforward neural network composed of an input layer, a set of hidden layers, and the output layer.

%Uma MLP mapeia uma função do tipo $f: x \rightarrow y$, onde $x$ são as características extraídas pelas camadas de convolução e \textit{pooling} e $y$ é a saída desejada da rede, ao executar um processo iterativo de ajustes de parâmetros (pesos) da rede MLP, normalmente utilizando o algoritmo \textit{Backpropagation} \citep{goodfellow2016deep}. Esse processo, conhecido como treinamento de uma ANN, é um problema de otimização que busca minimizar uma função de perda \citep{goodfellow2016deep}. A função de perda calcula o erro entre as saídas desejadas e as obtidas pela rede neural. O erro obtido é utilizado para recalcular os pesos da rede \citep{zhang2020dive}. A seguir é descrita a função de perda Entropia Cruzada, comumente utilizada por CNN's em problemas de classificação:
A MLP maps a $f: x \rightarrow y$ function, where $x$ are the features extracted by the convolution and pooling layers, and $y$ is the desired network output. This process, known as training an ANN, is an optimization problem that seeks to minimize a loss function \citep{goodfellow2016deep}. The loss function calculates the error between the desired outputs and those obtained by the neural network. The error obtained is used to recalculate the weights of the network \citep{zhang2020dive}. The Cross-Entropy loss function commonly used by CNN's in classification problems is described by:

\begin{equation}
    Cross\-Entropy(pa, pb) = - \sum_{i}^k pa_{i}\log{pb_{i}}
\end{equation}
where $pa_{i}$ and $pb_{i}$ represent the real and predicted probability for the class $i$, respectively and $k$ represents the number of classes. For unbalanced datasets, loss functions that consider the frequency of each class in its calculation, such as the Weighted Cross Entropy are often used.

Along with the parameter adjustment algorithm, weight regularization techniques can be included in order to avoid problems such as overfitting, when a model presents good results in the training process but it is not able to work properly to unknown data. In this work, the dropout regularization technique was applied. This technique consists of removing some of the neuron connections with a certain probability, in order to improve the network's generalization to unknown data \citep{zhang2020dive}. 

\begin{figure}[h!]
    \centering
    \includegraphics[width=1\textwidth]{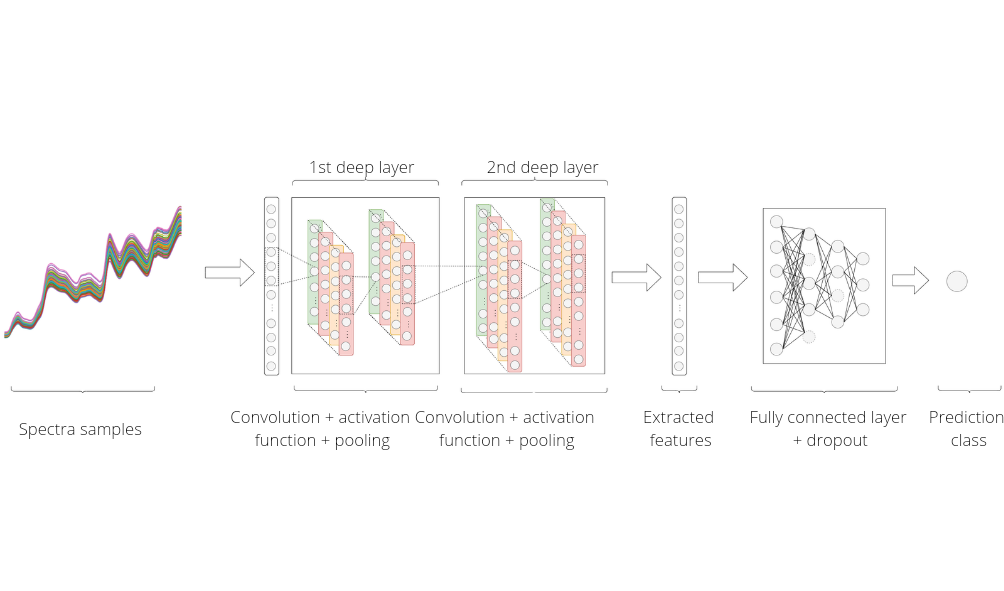} 
    \caption{1D-CNN architecture made up of convolution, pooling, and fully connected layers.}
    \label{fig-cnn}
\end{figure}

By combining multiple convolution and pooling layers with a fully connected layer is generated CNN-like architectures. Figure \ref{fig-cnn} shows the 1D-CNN architecture used to classify spectral data in this work. %Dados originados a partir de espectrômetros alimentam a rede, passando primeiramente por duas camadas de convolução + \textit{pooling}. A primeira camada irá produzir um mapa de características a partir dos dados de entrada. O mapa de características gerado servirá como entrada da segunda camada de convolução + \textit{pooling}, gerando um novo mapa de características, o qual servirá como entrada da camada totalmente conectada que irá gerar a saída da rede. A técnica de regularização por \textit{dropout} é representada na camada totalmente conectada por neurônios que possuem conexões removidas com as camadas posteriores a ele.

%%% Results
\section{Experiments and Results}
This section presents the results with discussions. First, the datasets used in the experiments are presented. Second, the algorithms are applied to the datasets in order to compare with machine learning and standard algorithms of chemometrics. Finally, a case study with spectral data from samples of the Sars-Cov-2 virus is presented.

\subsection{Datasets}
Next, we present a brief description of the datasets. \cite{kosmowski2018evaluation} collected infrared spectral data from barley, chickpea, and sorghum crops. Each type of grain originated a dataset. The Barley dataset is composed of 1200 barley grain samples, distributed in 24 classes of grain variants. The Chickpea dataset has 950 chickpea samples distributed in 19 classes, and finally, the Sorghum dataset has 500 sorghum samples distributed in 10 classes. Despite a large number of samples in the dataset, it is clear that they are distributed in a large number of classes. Consequently, there is a small frequency of samples per class. All three datasets are well balanced. %Na Figura \ref{dataset-grains} é mostrado, respectivamente, os espectro coletados para as bases de dados \textit{Barley}, \textit{Chickpea}, \textit{Sorghum}, com seus respectivos comprimentos de onda em nanômetros e níveis de absorção.
%\begin{figure}[h!]
%  \centering
%  \subfloat[Níveis de absorção para cada com-\\primento de onda para o espectro\\ da base de dados \textit{Barley}.]{\includegraphics[width=0.45\linewidth]{figuras/spectra_plots/barley_spectra.png}}
%  \subfloat[Níveis de absorção para cada comprimento de onda para o espectro da base de dados \textit{Chickpea}.]{\includegraphics[width=0.45\linewidth]{figuras/spectra_plots/chickpea_spectra.png}}\\
%  \subfloat[Níveis de absorção para cada comprimento de onda para o espectro da base de dados \textit{Sorghum}.]{\includegraphics[width=0.45\linewidth]{figuras/spectra_plots/sorghum_spectra.png}}
%  \caption{Espectro para as bases de dados \textit{Barley}, \textit{Chickpea} e \textit{Sorghum}.}
%  \label{dataset-grains}
%\end{figure}
\cite{zheng2019spectra} presented an approach that uses Extreme Learning Machine for spectra classification and made the data available. Four datasets were used in this study, Coffee, Meat, Oil, and Fruits. The Coffee dataset presents samples divided into Arabica and Conilon coffee beans uniformly distributed in 56 samples. The Meat dataset is made up of 60 samples of chicken, pork, and turkey, also evenly distributed. The Oil dataset contains 60 extra virgin olive oil spectrum samples from 4 different countries: Greece, Italy, Portugal, and Spain. The Fruits dataset consists of 983 samples of fruit purees divided into two classes, strawberry, and non-strawberry. %Na Figura \ref{dataset-coffee-meat} é mostrado os espectros coletados para as bases de dados \textit{Coffee} e \textit{Meat}. Na Figura \ref{dataset-oil-fruits} são apresentados os espectros coletados para as base de dados \textit{Oil} e \textit{Fruits}.
Table \ref{tab-base-de-dados} lists the number of samples and classes present in each dataset. 

%\begin{figure}
%  \begin{subfigure}[httb!]{0.5\textwidth}
%    \includegraphics[width=\textwidth]{figuras/spectra_plots/coffee_espectra.png}
%    \caption{Níveis de absorção para cada comprimento de onda para o espectro da base de dados \textit{Coffee}.}
%  \end{subfigure}
  %
%  \begin{subfigure}[httb!]{0.5\textwidth}
%    \includegraphics[width=\textwidth]{figuras/spectra_plots/meat_spectra.png}
%    \caption{Níveis de absorção para cada comprimento de onda para o espectro da base de dados \textit{Meat}.}
%  \end{subfigure}
%    \caption{Espectro para as bases de dados \textit{Coffee} e \textit{Meat}.}
%  \label{dataset-coffee-meat}
%\end{figure}

%\begin{figure}[httb!]
%  \begin{subfigure}[httb!]{0.5\textwidth}
%    \includegraphics[width=\textwidth]{figuras/spectra_plots/oliver_spectra.png}
%    \caption{Níveis de absorção para cada comprimento de onda para o espectro da base de dados \textit{Oil}.}
%  \end{subfigure}
  %
%  \begin{subfigure}[httb!]{0.5\textwidth}
%    \includegraphics[width=\textwidth]{figuras/spectra_plots/fruits_spectra.png}
%    \caption{Níveis de absorção para cada comprimento de onda para o espectro da base de dados \textit{Fruits}.}
%  \end{subfigure}
%    \caption{Espectro para as bases de dados \textit{Oil} e \textit{Fruits}.}
%  \label{dataset-oil-fruits}
%\end{figure}

\begin{table}[]
\centering
\begin{tabular}{ccc}
\hline
\multicolumn{1}{l}{Dataset} & \multicolumn{1}{l}{Samples} & \multicolumn{1}{l}{N. classes} \\ \hline
Barley                              & 1200                          & 24                              \\
Chickpea                            & 950                           & 19                              \\
Sorghum                             & 500                           & 10                              \\
Coffee                              & 56                            & 2                               \\
Meat                                & 120                           & 3                               \\
Oil                                 & 120                           & 4                               \\
Fruits                              & 983                           & 2                              
\end{tabular}
\caption{Number of samples and classes present in each dataset.}
\label{tab-base-de-dados}
\end{table}

\subsection{Experimental Setting}

The training  of 1-D CNN  and machine learning algorithms investigated in this work were performed using cross-validation. This method is used to decrease the bias and avoid over-estimated models. The learning process consists of dividing the dataset into $k$ \textit{folds} and use each one separately as a test set of a model. At the end, one calculates the average accuracy and standard deviation of the algorithm for the dataset. The process of training and evaluating an algorithm using \textit{cross-validation} is described in the Algorithm \ref{alg-cross-validation}. For all experiments a number of $k=5$ folds was adopted, in other words, the dataset was divided into $80\%$ for training and validation and $20\%$ was used exclusively for testing.

\begin{algorithm}[httb!]
\SetAlgoLined
    1 - Shuffle the samples.\\
    2 - Split it into $k$ folds. \\
    \For{f in folds }{
        1 - Keep $f$ as a test set.\\
        2 - Train a new model with the remaining folds\\
        3 - Evaluate the model with \textit{fold} $f$.\\
        4 - Save the model accuracy.\\
    }
    3 - Get the cross-validation mean accuracy and standard deviation.\\
    4 - Return the model within the closest accuracy to the cross-validation mean accuracy.
    \caption{Cross-validation evaluation}
    \label{alg-cross-validation}
 
\end{algorithm}

The algorithms PLS-DA, KNN, SVM, and CNN need hyperparameter adjustments. So, one can select the best configuration for each dataset and maximize its performance.

The PLS-DA algorithm is based on components construction in order to maximize the variance of the variables. Each component explains the degree of variance it represents. For each dataset,  $c$ components were selected that, together, represent 95\% of the total variance of the data.

For the KNN and SVM algorithms, a variation of the algorithm \ref{alg-cross-validation}  was used, in order to find the best set of hyperparameters for each cross-validation round. The data used to train the model undergoes a new division into a training set and validation set, in order to find the best combination of hyperparameters. Thus, for each combination of training and testing folds, one finds a set of optimal hyperparameters for the model. At the end of the process, one calculates the mean and standard deviation obtained by the algorithm and the set of hyperparameters that generated the model with the closest accuracy to the average accuracy, avoiding an overestimated model. The process for finding the best hyperparameters and average accuracy is described in the algorithm \ref{alg-cross-validation-2}.

Table \ref{tab-espaco-busca-pls-knn} presents the search space used for the KNN and SVM algorithms. Table \ref{tab-hiper-pls-knn} presents the number of components $c$ needed to obtain $95\%$ of the data variance in the PLS-DA algorithm, in addition to the amount of $k$-neighbors and the kernel function that provided the closest result to the average result obtained when executing the algorithm \ref{alg-cross-validation-2}.

\begin{table}[httb!]
\centering
\begin{tabular}{c|c|c}
\hline
Algorithm  & Hyperparameter & Search Space               \\ \hline
KNN       & \textit{k}-neighbors     & {[}2, 3, 5, 10, 15, 20, 24{]} \\ 
SVM       &  Kernel Function     & {[}Linear, Polinomial, Sigmoid, RBF{]} \\
\end{tabular}
\caption{Search space for KNN and SVM hyperparameters. }
\label{tab-espaco-busca-pls-knn}
\end{table}

\begin{algorithm}[httb!]
\SetAlgoLined
    1 - Shuffle the samples.\\
    2 - Split it into $k$ folds. \\
    \For{f in folds }{
        1 - Keep $f$ as a test set.\\
        \For{each set $\lambda$ of hyperparameters}{
            \For{ v in (folds - f)}{
                1 - Keep $v$ as a validation set.\\
                2 - Train a new model with the remaining folds.\\
                3 - Evaluate the model with fold $v$.\\
                4 - Save the model accuracy.\\
            } 
            1 - Save the mean accuracy and standard deviation for the $\lambda$ set. \\
        }
        2 - Train a new model with the remaining folds with the best $\lambda$ set.\\
        3 - Evaluate the model with fold $f$.\\
        4 - Save the accuracy and the model\\
    }
    3 - Get the cross-validation mean accuracy and standard deviation.\\
    4 - Return the model and $\lambda$ set within the closest accuracy to the cross-validation mean accuracy.
    \caption{Cross-validation evaluation - variation to find best set of hyperparameters}
    \label{alg-cross-validation-2}
 
\end{algorithm}

\begin{table}[httb!]
\centering
\begin{tabular}{c|c|c|c}
\hline
\multicolumn{1}{l|}{Datasets} & \multicolumn{1}{l|}{N. of components} & \multicolumn{1}{l|}{N. \textit{k}-neighbors} & \multicolumn{1}{l}{Kernel Function} \\ \hline
Barley                              & 2                                   & 15  &   Sigmoid                                \\
Chickpea                            & 2                                   & 10  &   Sigmoid                               \\
Sorghum                             & 2                                   & 10  &   Sigmoid                            \\
Coffee                              & 2                                   & 3   &   RBF                         \\
Meat                                & 3                                   & 3   &   RBF                       \\
Oil                                 & 6                                   & 5   &   RBF                       \\
Fruits                              & 4                                   & 3   &   RBF    
\end{tabular}
\caption{Hyperparameters that obtained the best results for the PLS-DA, KNN, and SVM.}
\label{tab-hiper-pls-knn}
\end{table}

Finally, the 1D-CNN, as a recent approach in the field of Chemometrics, it does not have pre-defined and pre-trained architectures for applications that involve spectral data classification. To choose the architecture and hyperparameters used in this work, the software Optuna\footnote{https://optuna.readthedocs.io/} was used, a hyperparameter optimizer developed to be used in machine learning problems that search for combinations of hyperparameters and architectures that maximize model performance. Similar to the algorithm \ref{alg-cross-validation-2}, it searches for the best model in the training and validation dataset and, after finding a set of suboptimal hyperparameters, tests the model in the test set. %Nos experimentos realizados foi utilizado uma arquitetura CNN-1D semelhante a apresentada na Figura \ref{fig-cnn}. A arquitetura é composta por três blocos principais: O primeiro e segundo blocos são formados ao combinar uma camada de convolução com uma camada de \textit{pooling}. O segundo bloco conecta-se com o terceiro, composto por uma camada totalmente conectada. Na Figura \ref{arq-base} é mostrada uma representação esquemática da arquitetura da rede CNN utilizada.

Table \ref{tab-hiperparam-cnn} lists the selected hyperparameters for each dataset involving the training and testing step. All experiments and algorithms were programmed using utilities from the scikit-learn\footnote{https://scikit-learn.org/} and PyTorch\footnote{https://pytorch.org/} libraries from Python\footnote{https: //www.python.org/}. The source code in the experiments is available from authors upon request.
%\begin{figure}[h!]
%    \centering
%    \includegraphics[width=\textwidth]{figuras/arq_base.png}
%    \caption{Composição da arquitetura da rede CNN utilizada nos experimentos.}
%    \label{arq-base}
%\end{figure}

\begin{table}[h!]
 \begin{adjustbox}{max width=\textwidth}
\begin{tabular}{c|c|c|c|c|c|c|c}
\cline{2-8}
                                            & \multicolumn{7}{c|}{Dataset}                                                                                                                                                                  \\ \hline
\multicolumn{1}{c|}{Hyperparameters}         & Barley                & Chickpea              & \multicolumn{1}{l|}{Sorghum} & \multicolumn{1}{l|}{Coffee} & \multicolumn{1}{l|}{Meat} & \multicolumn{1}{l|}{Oil} & \multicolumn{1}{l|}{Fuit} \\ \hline
\multicolumn{1}{c|}{Optimizer}            & Adam                  & Adam                  & Adam                         & Adam                        & Adam                      & Adam                     & Adam                      \\
\multicolumn{1}{c|}{Learning rate}   & 0.001                 & 0.001                 & 0.001                        & 0.0001                      & 0.001                     & 0.001                    & 0.001                     \\
\multicolumn{1}{c|}{Epochs}                & 5000                  & 5000                  & 5000                         & 100                         & 700                       & 1000                     & 1000                      \\
\multicolumn{1}{c|}{Dropout Rate}       & 0.4                   & 0.4                   & 0.3                          & 0.1                         & 0.1                       & 0.2                      & 0.1                       \\
\multicolumn{1}{c|}{Batch Size}       & 50                   & 50                   & 50                          & 6                         & 10                       & 10                      & 50                       \\
\multicolumn{1}{c|}{Loss Function}       & Cross entropy                   & Cross entropy                   & Cross entropy                          & Cross entropy                         & Cross entropy                       & Cross entropy                      & Cross entropy                       \\
    
\end{tabular}
\end{adjustbox}
\caption{Hyperparameters that obtained the best results for 1D-CNN.}
\label{tab-hiperparam-cnn}
\end{table}

\newpage
\subsection{Experimental Results}
\label{sec-resultados}

This subsection presents the results obtained by each algorithm when applied to the datasets in this work. The experiments were split into two parts: 1) Using original spectral data, and 2) Applying the Savitzky–Golay (SG) filter as a pre-processing step. The division of the experiments into two stages seeks to analyze the influence of data pre-processing on the model results. All experiments were performed on a machine with Windows 10, equipped with an Intel Core I5-8265U processor and 8GB of RAM memory.

\paragraph{Original Data:} The results obtained for the first configuration of experiments, data without pre-processing, are presented in table \ref{tab-resultados-1}. The investigated algorithms (MOCHM, PLS-DA, SVM, KNN and, 1D-CNN) were applied to each dataset. Best results are shown in bold.

\begin{table}[h!]
\centering
\begin{tabular}{c|c|c|c|c|c}
\hline
Dataset  & MOCHM                           & PLS-DA                            & SVM                              & KNN                              & 1D-CNN                             \\ \hline
Barley   & -                               & $10.66 \pm 0.62$                      & $7.75 \pm 0.54$                      & $16.08 \pm 1.38$                     & \textbf{47.4 $\pm$ 4.7}                      \\
Chickpea & -                               & $14.73 \pm 1.12$                      & $23.58 \pm 1.07$                     & $27.16\pm4.48$                       & \textbf{54.2 $\pm$ 3.3}                      \\
Sorghum  & -                               & $18.79 \pm 3.31$                      & $15.79 \pm 1.94$                     & $34.79 \pm 2.99$                     & \textbf{55.8 $\pm$ 3.3}                      \\
Coffee   & $51 \pm 2.1$                      & $85.7 \pm 3.4$                        & $42.7 \pm 3.4$                       & $46.3 \pm1.37$                       & \textbf{100 $\pm$ 0}                             \\
Meat     & $36 \pm 0.2$                      & $70.83 \pm 6.56$                      & $41.6 \pm 3.1$                       & $87.5 \pm 7$                         & \textbf{95 $\pm$ 3.3}                        \\
Oil      & \multicolumn{1}{l|}{$44 \pm 0.7$}  & \multicolumn{1}{l|}{$75 \pm 2.04$}    & \multicolumn{1}{l|}{$41.6 \pm 1.02$} & \multicolumn{1}{l|}{$73.3 \pm 3.11$} & \multicolumn{1}{l}{\textbf{88.8 $\pm$ 5.3}} \\
Fruit    & \multicolumn{1}{l|}{$95 \pm 1.9$} & \multicolumn{1}{l|}{$91.04 \pm 1.02$} & \multicolumn{1}{l|}{$88.3\pm2.96$}   & \multicolumn{1}{l|}{$88.9 \pm1.75$}  & \multicolumn{1}{l}{\textbf{96 $\pm$ 0.6}}  
\end{tabular}
\caption{Average accuracy obtained by the algorithms when applied to data without pre-processing.}
\label{tab-resultados-1}
\end{table}

The obtained results  with the 1D-CNN were superior for all datasets using data without pre-processing. It is observed, however, that the datasets \textit{Barley}, \textit{Chickpea} and \textit{Sorghum} achieved an average accuracy lower than other datasets. It turns out that due to the high number of classes present in the data and the small number of samples available for each class in the training step, the model was not able to generalize effectively. Nevertheless, we notice a difference when comparing the results obtained by 1D-CNN with those obtained by the standard algorithms used in Chemometrics, i.e., PLS-DA, and SVM. The KNN algorithm also obtained better results than the PLS-DA and SVM algorithms, reaching an average accuracy of $20\%$ for the \textit{Sorghum} dataset compared to SVM, the worst algorithm performance at the dataset. In the other datasets, the PLS-DA algorithm obtained, in general, the best results after 1D-CNN. The MOCHM algorithm was stoped before finishing its execution for the \textit{Barley}, \textit{Sorghum} and \textit{Chickpea} datasets because it demanded a computation time over 60 minutes to finish the process.

%Na tabela \ref{tab-resultados-rank1} é apresentado o ranqueamento dos algoritmos por base de dados e o ranqueamento médio obtido. Como discutido anteriormente, a CNN obteve o melhor desempenho, seguidos por PLS-DA e KNN, que obtiveram o mesmo ranqueamento médio. 

%\begin{table}[h!]
%\centering
%\begin{tabular}{cccccc}
%\hline
%Dataset                     & MOCHM & PLS-DA & SVM  & KNN  & 1D-CNN \\ \hline
%Barley                      & 5     & 3      & 4    & 2    & 1   \\
%Chickpea                    & 5     & 4      & 2    & 3    & 1   \\
%Sorghum                     & 5     & 3      & 2    & 2    & 1   \\
%Coffee                      & 3     & 2      & 5    & 4    & 1   \\
%Meat                        & 5     & 3      & 4    & 2    & 1   \\
%Oil                         & 4     & 2      & 5    & 3    & 1   \\
%Fruit                       & 2     & 3      & 5    & 4    & 1   \\ \hline
%\multicolumn{1}{l}{Média} & 4.14  & 2.85   & 3.85 & 2.85 & 1   \\ \hline
%\end{tabular}
%\caption{Ranqueamento dos algoritmos por base de dados e ranqueamento médio obtido ao utilizar dados sem pré-processamento.}
%\label{tab-resultados-rank1}
%\end{table}

 \paragraph{Pre-processed data:} The second configuration of experiments was performed with the same algorithms and datasets, but applying the SG filter in the pre-processing step. The SG filter was performed with a standard hyperparameter configuration: the window size was set to 11, using a polynomial of degree 3 and second-order derivative. In the table \ref{tab-resultados-sg} the results obtained in the experiment are presented.

\begin{table}[h!]
\centering
\begin{tabular}{c|c|c|c|c|c}
\hline
Dataset  & MOCHM                 & PLS-DA                           & SVM                               & KNN                               & 1D-CNN                                       \\ \hline
Barley   & -                     & 11.83 $\pm$ 2.77                     & 14.25 $\pm$ 2.3                       & \textbf{64.91 $\pm$ 1.3}              & 46.16 $\pm$  0.97                             \\
Chickpea & -                     & 15.78 $\pm$ 1.07                     & 20.4 $\pm$ 4.28                       & \textbf{71.36 $\pm$ 2.11}             & 55.2 $\pm$ 1.1                                 \\
Sorghum  & -                     & 25.99 $\pm$ 2.63                     & 28.38 $\pm$ 5.87                      & \textbf{71.39 $\pm$ 3.3}              & 59.6 $\pm$ 2.1                                \\
Coffee   & 83 $\pm$ 1.07                      & 87.4 $\pm$ 2.1                       & 92.69 $\pm$ 6.1                       & 80.11 $\pm$ 4.9                       & \textbf{100 $\pm$ 0}                          \\
Meat     & 81 $\pm$ 0.8                      & 71.66 $\pm$ 4.24                     & 58.33 $\pm$ 3.1                       & 92.49 $\pm$ 4.08                      & \textbf{96.6 $\pm$ 2.07}                      \\
Oil      & \multicolumn{1}{l|}{ \textbf{91 $\pm$ 1.1 }} & \multicolumn{1}{l|}{75.8 $\pm$ 3.11} & \multicolumn{1}{l|}{41.6 $\pm$ 1.02}  & \multicolumn{1}{l|}{79.1 $\pm$ 1.16}  & \multicolumn{1}{l}{89.4 $\pm$ 1.12} \\
Fruit    & \multicolumn{1}{l|}{ 92 $\pm$ 2.2} & \multicolumn{1}{l|}{79.2 $\pm$ 1.02}  & \multicolumn{1}{l|}{93.49 $\pm$ 0.86} & \multicolumn{1}{l|}{91.86 $\pm$ 0.38} & \textbf{94.2 $\pm$ 0.3}                       
\end{tabular}
\caption{Average accuracy obtained by the algorithms when applied to data with pre-processing.}
\label{tab-resultados-sg}
\end{table}

We can observe that the 1D-CNN did not obtain the hegemony of the results as reported in the previous case. For the \textit{Barley}, \textit{Chickpea} and \textit{Sorghum} datasets, KNN was superior to CNN. For the other datasets, CNN maintained its superiority except for the \textit{Oil} dataset, where the MOCHM obtained the best accuracy.

In general, the performance gain of the algorithms when using pre-processed data is noticeable. We can evaluate the gain from two aspects: 1) The increase in the average accuracy of the algorithms for the datasets, with emphasis on the MOCHM and KNN algorithms, and 2) The reduction in standard deviation, as observed on 1D-CNN.

The MOCHM achieved a gain of approximately $50\%$ in accuracy in the \textit{Oil} dataset, achieving better overall performance. KNN also achieved significant gains in accuracy surpassing CNN in the \textit{Barley}, \textit{Chickpea} and \textit{Sorghum} datasets. Regarding the decrease in standard deviation, CNN presents this characteristic for all datasets. Despite not presenting a high gain in accuracy as in other algorithms, the reduction in the standard deviation indicates greater stability, generating models with a higher degree of reliability.  This fact is explicit in the results obtained by the 1D-CNN in the \textit{Barley} dataset, where it reached an accuracy of $47.4 \pm 4.7$ in the experiments without preprocessing against $46.16 \pm 0.97$ in the experiments with preprocessing. 

%Na tabela \ref{tab-resultados-rank2} está listado o ranqueamento dos algoritmos por base de dados e seu respectivo ranqueamento médio. Novamente a CNN obteve o melhor desempenho geral. No entanto ao utilizar dados pré-processados o algoritmo KNN obteve superioridade ao PLS-DA, o que não havia ocorrido anteriormente como apresentado na tabela \ref{tab-resultados-rank1}. Os algoritmo PLS-DA e SVM obtiveram mesmo ranqueamento médio, seguidos pelo algoritmo MOCHM.

%\begin{table}[h!]
%\centering
%\begin{tabular}{cccccc}
%\hline
%Dataset                     & MOCHM & PLS-DA & SVM  & KNN  & CNN  \\ \hline
%Barley                      & 5     & 4      & 3    & 1    & 2    \\
%Chickpea                    & 5     & 4      & 3    & 1    & 2    \\
%Sorghum                     & 5     & 3      & 4    & 1    & 2    \\
%Coffee                      & 4     & 2      & 4    & 5    & 1    \\
%Meat                        & 3     & 4      & 5    & 2    & 1    \\
%Oil                         & 1     & 4      & 5    & 2    & 2    \\
%Fruit                       & 4     & 5      & 2    & 3    & 1    \\ \hline
%\multicolumn{1}{l}{Média} & 3.85  & 3.71   & 3.71 & 2.14 & 1.57 \\ \hline
%\end{tabular}
%\caption{Ranqueamento dos algoritmos por base de dados e ranqueamento médio obtido por cada algoritmo ao utilizar dados pré-processados.}
%\label{tab-resultados-rank2}
%\end{table}

Despite the superior results obtained by 1D-CNN in terms of accuracy  for the experiments performed, it is necessary to analyze the trade-off in the results. As 1D-CNN is an algorithm which demands a higher computational cost compared to other algorithms, then it is necessary to analyze in which cases its use is preferred. Table \ref{tab-tempo-treinamento} presents the training time required for a model of each algorithm per dataset. It is clear that the larger the dataset, the longer the time needed for training the 1D-CNN. Furthermore, it was the only algorithm that needs, in most cases, minutes to train the model. In problems with a large number of samples and high dimensionality, the efficient use of pre-processing algorithms combined with simple algorithms, such as KNN, might be a viable alternative instead of using 1D-CNN.

\begin{table}[h!]
\centering
\begin{tabular}{c|c|c|c|c|c}
\hline
Dataset  & MOCHM         & PLS-DA & SVM    & KNN    & 1D-CNN \\ \hline
Barley   &   -           & 0.1328 & 1.7173 & 0.3353 & 603.12    \\
Chickpea &   -           & 0.0855 & 0.8775 & 0.2436 & 485.30    \\
Sorghum  &   -   	     & 0.0255 & 0.2826 & 0.1020 & 443.05    \\
Coffee   &   22.10       & 0.1196 & 0.0083 & 0.0795 & 4.22    \\
Meat     &   18.30       & 0.0377 & 0.0136 & 0.0277 & 45.27    \\
Oil      &   31.55       & 0.0152 & 0.0194 & 0.0231 & 106.07    \\
Fruit    &   32.26       & 0.0377 & 0.3920 & 0.2325 & 81.13   
\end{tabular}
\caption{Model Training time for each algorithm in seconds.}
\label{tab-tempo-treinamento}
\end{table}

\paragraph {Data visualization:} In order to get a 2D visualization of the data pre-processing influence, the t-Distributed Stochastic Neighbor Embedding (t-SNE) \citep{van2008visualizing} was applied to the cases without and with pre-processing.

\begin{figure}[httb!]
  \begin{subfigure}[httb!]{0.5\textwidth}
    \includegraphics[width=\textwidth]{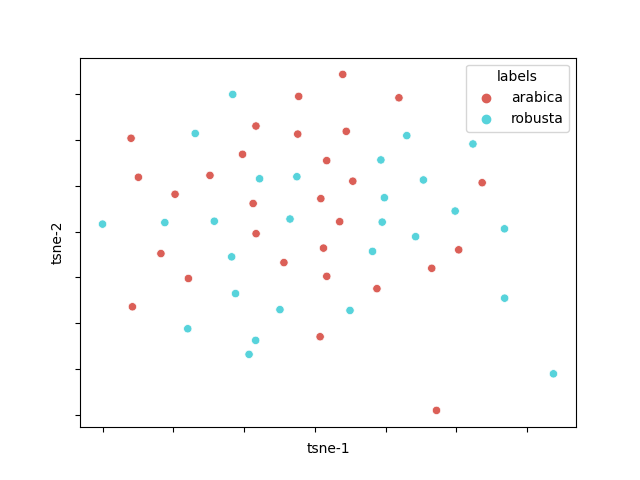}
    \caption{}
  \end{subfigure}
  \begin{subfigure}[httb!]{0.5\textwidth}
    \includegraphics[width=\textwidth]{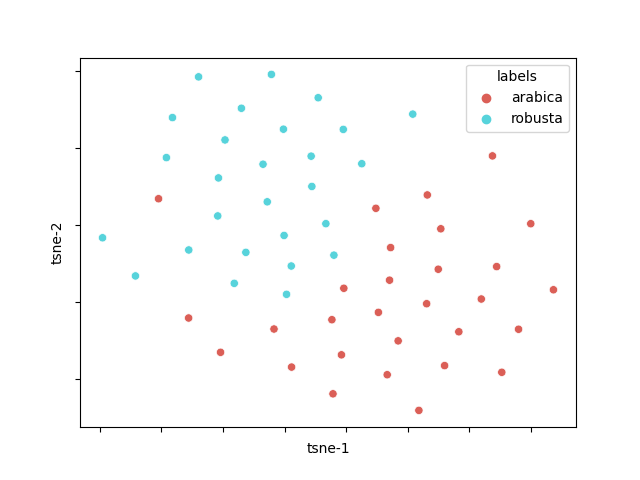}
    \caption{}
  \end{subfigure}
    \caption{Visualization of the spatial arrangement of (a) original data and (b) data after pre-processing using t-SNE for the Coffee dataset.}
  \label{feature-coffee}
\end{figure}

Figure \ref{feature-coffee} shows the original data and with pre-processing for the \textit{Coffee} dataset. It is not possible to observe in the original data clear groups that distinguish the two classes present in the dataset so that a simple classifier had difficulties to correctly separate the samples through a linear function. However, looking at the pre-processed data, we see the data grouped in a concise way, showing that the SG filter helps the model to discriminate the data.
%These data corroborate the results presented in the tables \ref{tab-resultados-1} and \ref{tab-resultados-sg}, where the \textit{Coffee} dataset obtained a higher accuracy we look at algorithms that do not have feature extractors, such as KNN and MOCHM.

%Na figura \ref{feature-sorghum} é mostrado a visualização de dados para a base \textit{Sorghum}. Esta base obteve resultados ruins ao utilizar dados originais, com exceção dos resultados obtidos pela CNN, que possui mecanismo de extração de características. Ao observamos a visualização dos dados sem pré-processamento, percebemos uma taxa elevada de superposição das amostras além de não ser possível distinguir grupos bem formados. Tais características podem ter influenciado no resultado de algoritmos que não possuem mecanismo de extração de características. Por outro lado, a figura referente a visualização de dados pré-processados começam a apresentar pequenos grupos, embora ainda haja superposição de amostras em alguns casos, o que ajuda a explicar a dificuldade dos algoritmos em obter uma acurácia elevada para essa base. 

%\begin{figure}[httb!]
%\centering
%  \begin{subfigure}[httb!]{0.7\textwidth}
%    \includegraphics[width=\textwidth]{figuras/feature_visualization/shorgum_raw.png}
%    \caption{}
%  \end{subfigure}
  %
%  \begin{subfigure}[httb!]{0.7\textwidth}
%    \includegraphics[width=\textwidth]{figuras/feature_visualization/sorghum_filter.png}
%    \caption{}
%  \end{subfigure}
%    \caption{Visualização da disposição espacial dos dados (a) originais e (b) após pré-processamento utilizando t-SNE para a base de dados \textit{Sorghum}.}
%  \label{feature-sorghum}
%\end{figure}

The results show the importance of applying pre-processing algorithms to spectral data for noise reduction and how this positively impacts the results obtained by each algorithm, whether in terms of accuracy gain or model stability. Convolutional Neural Networks are promising in the analysis of spectral data, since the results obtained indicate superior performance over traditional algorithms. It is worth to mention the ability of 1D-CNN to generalize well for small datasets. Furthermore, algorithms such as KNN prove to be a viable and simple alternative to use in most cases when combined with a data pre-processing algorithm, since it obtains good results and does not demand high computational costs. Based on the results obtained, we extend our work to a case study using 1D-CNN, PLS-DA, and KNN in the next section, since these algorithms obtained the best average performance in the experiments previously investigated.

\subsection{Covid-19 case study}

The Sars-Cov-2 virus (COVID-19) pandemic has emerged as the biggest and challenging  health problem facing the world in recent decades. With the first  case confirmed in December, 2019 in Wuhan District, China \citep{zhu2020novel}, the virus has spread rapidly in China and to other countries, and in January 2020 the World Health Organization (WHO) declared a state of global emergency  \citep{WHO2019_COVID}. 

Due to its rapid transmission characteristics and high transmission rate, in a few months since the first confirmed case, the virus has been reported in 144 countries around the world \citep{world2020world}. In order to carry out sanitary restrictions, assess and report on the spread of the virus, mass testing was started in several territories that had cases of COVID-19. However, the available methods for testing have undesirable characteristics that can influence the decisions taken to contain the virus. First, the high cost of available tests, causing the unavailability of testing in underdeveloped countries. Second, the time needed to obtain definitive results from the presence of the virus in the patient. Third, the high rate of false-negative tests. Among the mentioned problems, the high rate of false negatives is a crucial factor that has a direct impact on the virus containment. Individuals tested as false negatives can relax the sanitary measures imposed by healthy organizations and increase the spread of the virus \citep{west2020covid}. Recently, there is a growing research interest to automated diagnose of COVID-19 using artificial intelligence \citep{khana2021ESWA}.

In this context, automated diagnostic applications using data from spectrometers may become an alternative to traditional tests. Since spectrometers are portable and easy-to-use device, the data collection performed by such devices in conjunction with computational classification models might provide a quick diagnosis in minutes. Such characteristics would affect positively the impact of two of the undesirable characteristics on current testing, high cost, and time required for the result. Finally, in order to reduce false-negative results, the third problem found in current testing, two additional metrics will be included in the analysis of results along with accuracy (ACC) i.e., specificity (ESPEC) and sensitivity (SE), common metrics in biomedical applications.

The specificity is calculated by:

\begin{equation}\label{eq-precision}
    ESPEC = \frac{TN}{TN + FP}
\end{equation}
where $TN$ represents the true negative values, i.e, the samples that belonged to the control group and the classifier classified them correctly, and $FP$ the false-positive values, representing the control group samples that were classified positively for Covid-19. 

The sensitivity metric (also known as recall) is calculated by:

\begin{equation}\label{eq-recall}
    SE = \frac{TP}{TP + FN}
\end{equation}
where $TP$ represents the true-positive values, i.e, the virus samples that were correctly classified, and $FN$ represents the false negative values, i.e, the samples that belonged to the Covid-19 group but were classified as a control group.

By using such metrics, we can answer two important questions: 
\begin{itemize}
    \item Among those samples labeled as a control group, in this case, samples that did not carry COVID-19, how many did the model correctly classify?
    \item For all the samples that are positive for COVID-19, how many did the model classify as carrying the virus? 
\end{itemize}
Especially in case of COVID-19, due to its high dispersion and lethality characteristics, we are interested in answering the second question, as this information is directly linked to the sensitivity metric. By obtaining a high sensitivity rate in the model, we are ensuring that cases in which COVID-19 is present in the sample are classified as positive for the virus, i.e, it reduces the number of false-negative diagnoses, a current problem faced in the test protocols used for detecting the virus.

\begin{figure}[h!]
    \centering
    \includegraphics[width=\textwidth]{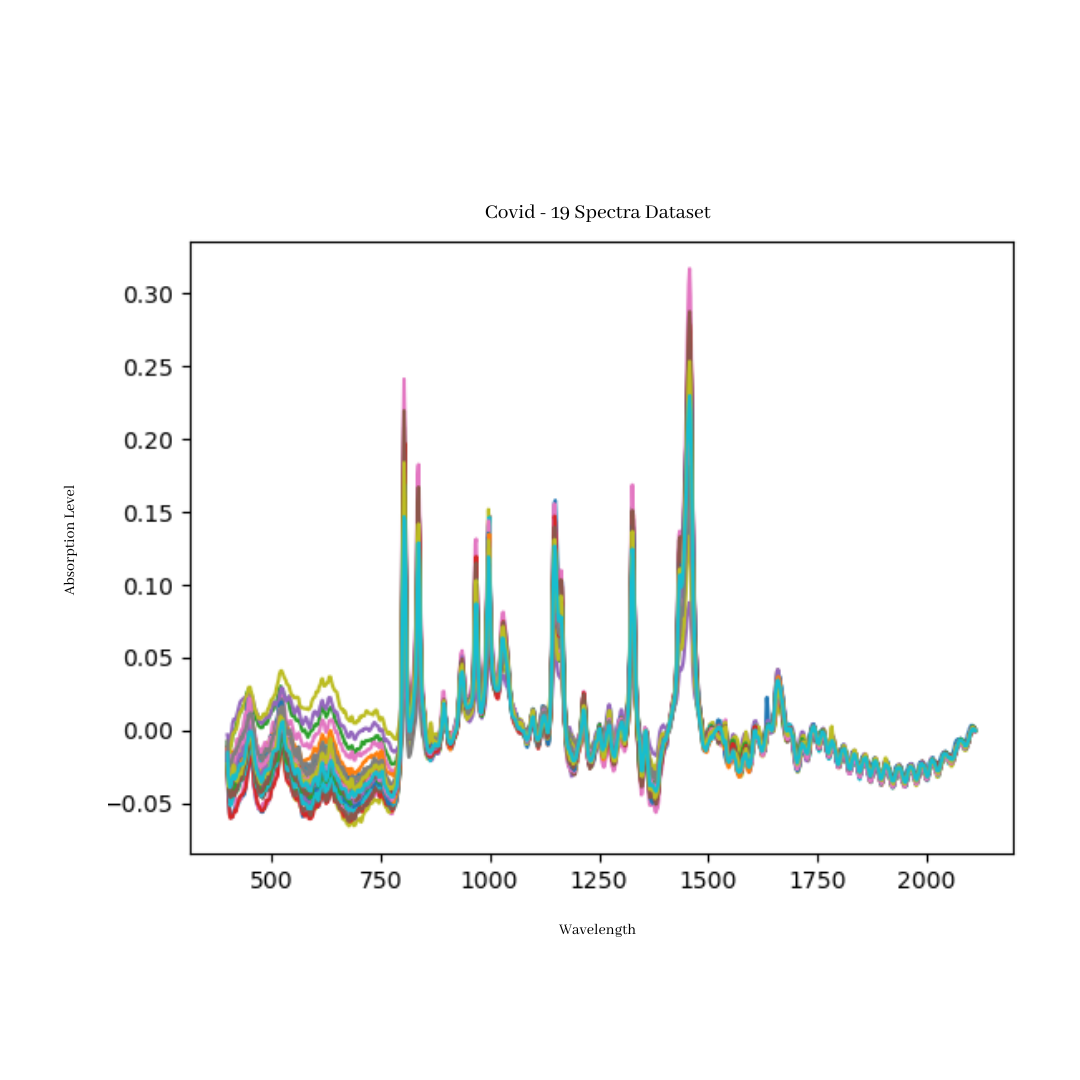}
    \caption{Absorption levels for each wavelength in nanometers for the spectrum collected from the Covid-19 dataset \citep{yin2021efficient}.}
    \label{spectra-covid}
\end{figure}

\cite{yin2021efficient} collected spectral samples from the control group and from patients infected with COVID-19 and made them available for public use. The dataset provided includes 309 samples, of which 150 belong to the control group and 159 to the group of patients infected by the virus. The collected spectra is shown in Figure \ref{spectra-covid}.

Table \ref{tab-resultados-covid} presents accuracy, specificity, and sensitivity results obtained for each algorithm when applied to the \textit{Covid} dataset \citep{yin2021efficient}, with original data and pre-processed by the SG filter. Similar to the results obtained for other datasets reported in the previous section, the 1D-CNN obtained the best results with an average accuracy of $96.5 \pm 1.1$ when using the SG filter in the pre-processing step. Furthermore, it is important to highlight the high values of specificity and sensitivity, with an average of $98.06 \pm 1.86$ and $94.06 \pm 2.43$, respectively. As explained before, low sensitivity values indicate a high number of false negatives, considered the worst scenario in the diagnosis of COVID-19. Such behavior can be observed in the results obtained by the KNN, which obtained, on average, a rate higher than $25\%$ of false negatives for data without pre-processing and $12\%$ for pre-processed data. This value indicates the infeasibility of the algorithm for real applications in the context of COVID-19. The PLS-DA algorithm presented, in general, promising results for all the analyzed metrics, however inferior to those presented by the 1D-CNN.

\begin{table}[httb!]
\centering
\begin{tabular}{c|c|l|l}
\hline
COVID-19       & ACC                & \multicolumn{1}{c|}{ESPEC}                 & \multicolumn{1}{c}{SE}                    \\ \hline
1D-CNN         & 94.19 $\pm$ 1.6         & \multicolumn{1}{c|}{96.82 $\pm$ 2.3}           & \multicolumn{1}{c}{94.72 $\pm$ 3.39}          \\
SG + 1D-CNN    & \textbf{96.5 $\pm$ 1.1} & \multicolumn{1}{c|}{\textbf{98.06 $\pm$ 1.86}} & \multicolumn{1}{c}{\textbf{94.06 $\pm$ 2.43}} \\
PLS-DA      & 93.7 $\pm$ 1.81        & 92.85 $\pm$ 1.8                                & 90.3 $\pm$ 2.1                                                \\
SG + PLS-DA & 94.17 $\pm$ 0.7          & 95.12 $\pm$ 1.15                               & 92.1 $\pm$ 1.9                                               \\
KNN         & 83.17 $\pm$ 4.84        & 92.78 $\pm$ 7.1                                & 73.59 $\pm$ 10.1                               \\
SG + KNN    & 93.2 $\pm$ 1.58         & 97.23 $\pm$ 0.9                                & 88.71 $\pm$ 1.1                               
\end{tabular}
\caption{Results obtained for the Covid dataset.}
\label{tab-resultados-covid}
\end{table}

\begin{figure}[h!]
  \centering
  \subfloat[SG + 1D-CNN - Confusion matrix ]{\includegraphics[width=0.45\linewidth]{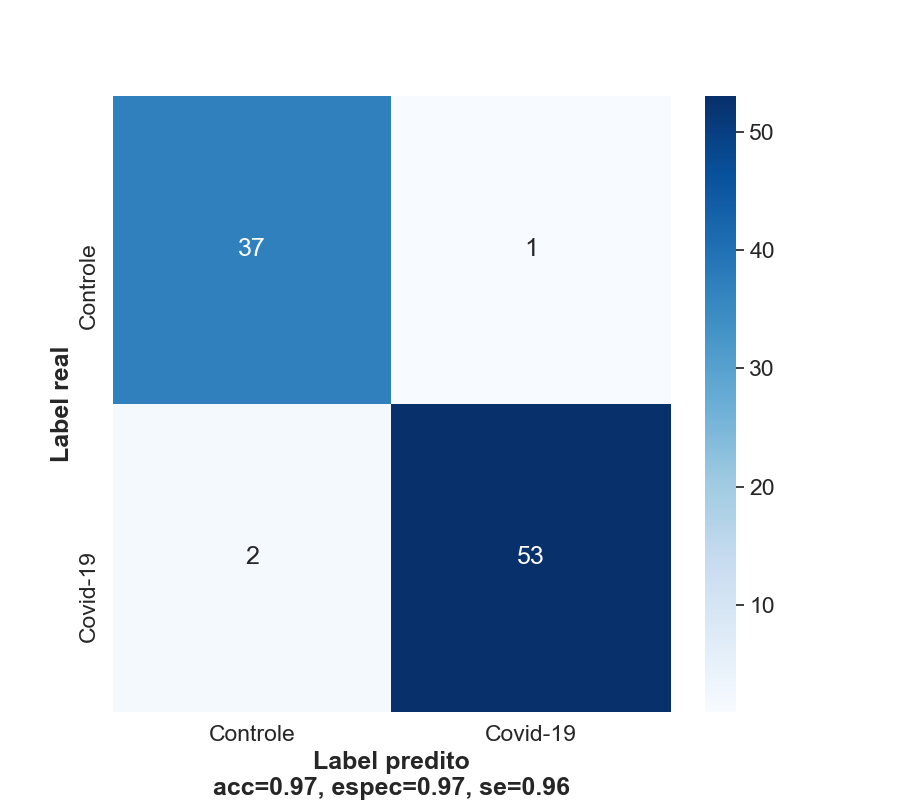}}
  \subfloat[SG + PLS-DA - - Confusion matrix]{\includegraphics[width=0.45\linewidth]{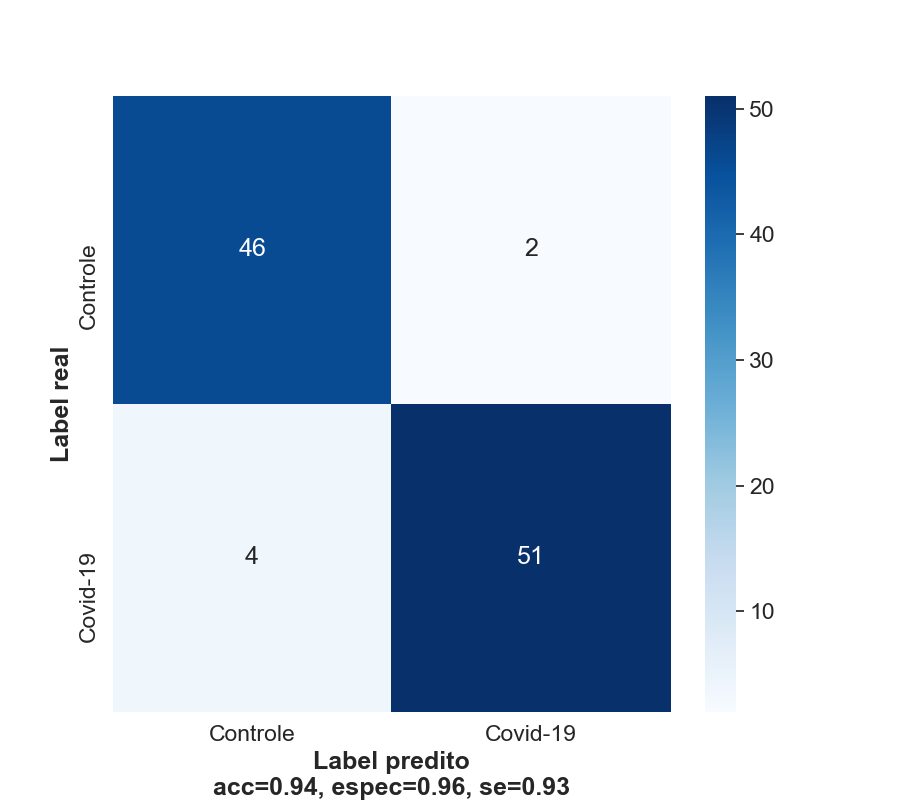}}\\
  \subfloat[SG + KNN - - Confusion matrix]{\includegraphics[width=0.45\linewidth]{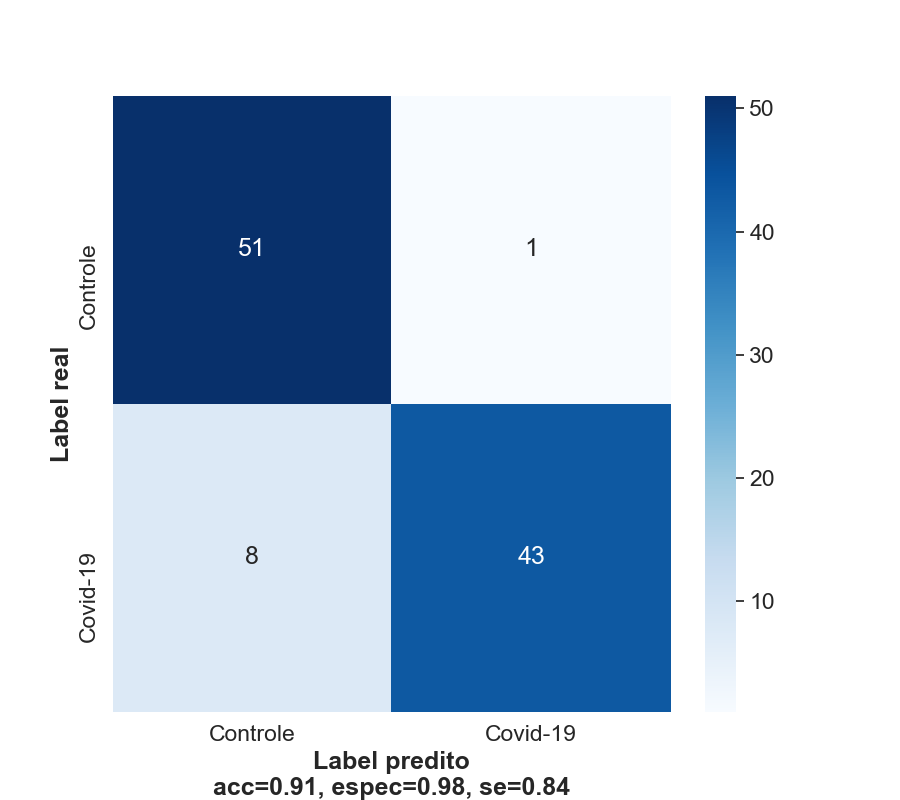}}
  \caption{Confusion matrices obtained by the models.}
  \label{fig-confusao}
\end{figure}

Figure \ref{fig-confusao} shows the confusion matrices for each algorithm using pre-processing. Each matrix is equivalent to the model that obtained the closest accuracy to the average accuracy obtained through the \textit{cross-validation} technique. The confusion matrix referring to 1D-CNN presents a high hit rate and only two false-negative, which confirms the results presented previously. The KNN confusion matrix presents a high amount of false negatives, in agreement with the results presented in table \ref{tab-resultados-covid}.

\begin{figure}[h!]
  \begin{subfigure}[httb!]{0.5\textwidth}
    \includegraphics[width=\textwidth]{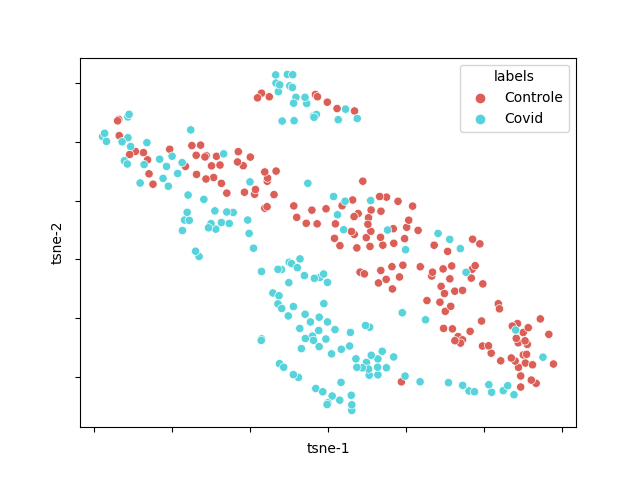}
    \caption{Spatial arrangement of original data.}
    \label{fig-a}
  \end{subfigure}
  \begin{subfigure}[httb!]{0.5\textwidth}
    \includegraphics[width=\textwidth]{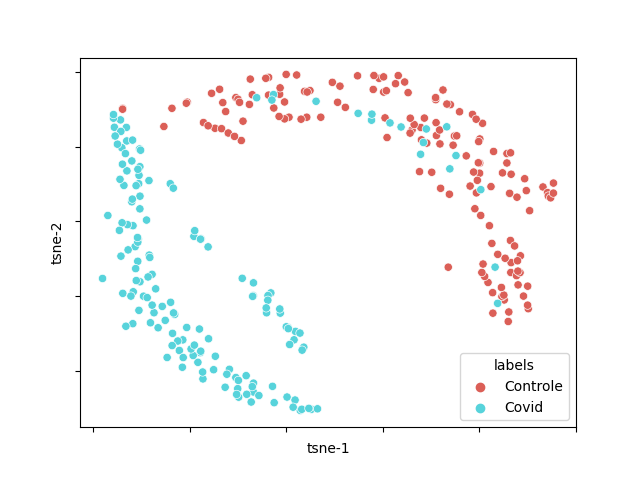}
    \caption{Spatial arrangement of pre-processed data.}
    \label{fig-b}
  \end{subfigure}
    \begin{subfigure}[httb!]{0.5\textwidth}
    \includegraphics[width=\textwidth]{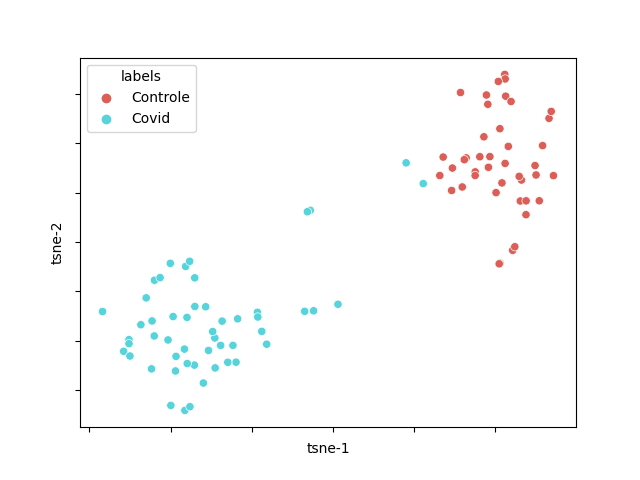}
    \caption{Spatial arrangement of features \\ extracted by 1D-CNN from original data.}
    \label{fig-c}
  \end{subfigure}
  \begin{subfigure}[httb!]{0.5\textwidth}
    \includegraphics[width=\textwidth]{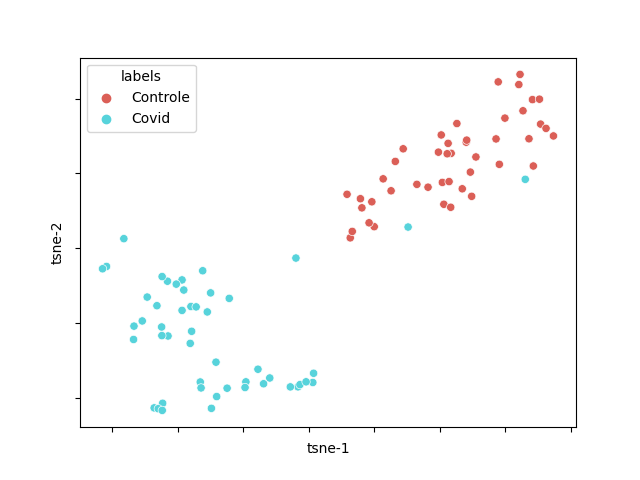}
    \caption{Spatial arrangement of features \\ extracted by 1D-CNN from pre-processed data.}
    \label{fig-d}
  \end{subfigure}
    \caption{Data spatial arrangement visualization using t-SNE for the Covid-19 dataset.}
  \label{feature-covid}
\end{figure}

The spatial arrangement of the data is shown in Figure \ref{feature-covid}. Figure \ref{fig-a} shows the visualization of the original data, Figure \ref{fig-b} the visualization of the data after pre-processing, and in Figures \ref{fig-c} and \ref{fig-d}  the features extracted by 1D-CNN for original and pre-processed data, respectively. Figure \ref{fig-a} shows that the original data presents overlapping and are not well grouped between the two classes. In Figure \ref{fig-b}, although the application of the SG filter improves the separation of the data into better defined groups, there is still an overlap of data in one of the groups, which makes the classification task difficult for algorithms that do not have feature extraction mechanisms like KNN. This problem can be solved by feature extractors, present on 1D-CNN. The visualization of the features extracted by   1D-CNN can be observed in Figures \ref{fig-c} and \ref{fig-d}, where we notice a better arrangement of groups than in those in Figures \ref{fig-a} and \ref{fig-b}. Especially, in Figure \ref{fig-c},  we note that 1D-CNN was able to group the data even without using the pre-processing technique.

Recent studies have applied spectroscopy techniques in conjunction with classification algorithms to analyze data collected from COVID-19 samples. \cite{yin2021efficient} applied the SVM algorithm to their dataset to perform the classification between the control group and the virus-infected group. \cite{Barauna2020} used ATR-FTIR spectroscopy to collect data from COVID-19 samples. In their study, it was used 111 samples for the control group and 70 positives for COVID-19. The authors applied a Genetic Algorithm with Linear Discriminant Analysis (GA-LDA)  to the data pre-processed by the SG filter to perform the classification task. \cite{carlomagno2021covid} also proposed the use of CNN-type networks for the classification of spectral samples from COVID-19, with a dataset containing 33 control samples and 30 positive samples for the virus. Table \ref{covid-final} presents the results obtained in this work and the works previously mentioned in the literature.

\begin{table}[httb!]
\centering
\begin{threeparttable}
\begin{tabular}{c|c|l|l}
\hline
Algoritmo   & ACC            & \multicolumn{1}{c|}{ESPEC}               & \multicolumn{1}{c}{SE}               \\ \hline
SG + 1D-CNN* & 96.5 $\pm$ 1.1 & \multicolumn{1}{c|}{98.06 $\pm$ 1.86} & \multicolumn{1}{c}{94.06 $\pm$ 2.43} \\
\cite{yin2021efficient}*    & {91 $\pm$ 4}               &                 93 $\pm$ 6                    & {89 $\pm$ 7}                                      \\
\cite{Barauna2020}**    & {90 $\pm$ 0}                &                      89 $\pm$ 0                &  {95 $\pm$ 0}                                      \\
\cite{carlomagno2021covid}**    &    {97.8 $\pm$ 0}            &               98 $\pm$ 0                     &    {97.5 $\pm$ 0}                                  
\end{tabular}
\begin{tablenotes}\footnotesize
\small
\item[*] Tests performed using the same database.
\item[**] Tests reporting author's own dataset and not yet publicly available.

\end{tablenotes}
\end{threeparttable}
\caption{Positioning of our results in respect to other studies in the literature for different spectral datasets obtained from Sars-Cov2 spectroscopy samples.}
\label{covid-final}
\end{table}

Although the dataset used was not the same for all studies, the approaches applied in this work and in the work of \cite{carlomagno2021covid} using        1D-CNN obtained superior results in respect to the metrics used. As noted earlier in the comparative study, steps such as data pre-processing can significantly influence the final classification results. It is worth mentioning the superiority of 1D-CNN in relation to SVM presented by \cite{yin2021efficient} using the same the Covid-19 dataset. The results presented by \cite{Barauna2020} using the GA-LDA algorithm show promising sensitivity values, surpassing those obtained in this work by $1\%$. However, they do not present standard deviation values. The results presented by \cite{carlomagno2021covid} and \cite{Barauna2020} may suffer changes when going through methods such as \textit{cross-validation}. In general, the results obtained using data pre-processed by the SG filter and 1D-CNN show promising results in terms of accuracy and sensitivity metrics for the dataset investigated, indicating the feasibility of automated systems to aid in the diagnosis of the COVID-19 using spectroscopy data.

%%% Conclusion
\section{Conclusion}

In this work, we present a CNN-1D to tackle classification of spectral data. Firstly, we carried out a comparative study between 1D-CNN and machine learning algorithms as SVM, PLS-DA, KNN and also MOCHM with the objective of evaluating the performance of these algorithms in chemometrics problems. We investigate the impact of data pre-processing and analyzed how this technique influences the performance of each algorithm. The results indicate that the 1D-CNN obtained an average performance superior to the other algorithms investigated when applied to spectroscopy datasets. Furthermore, data pre-processing proved to be a crucial step in the problem modeling process, influencing the gain on average accuracy and reduction of standard deviation, which impacts the reliability of the results. Besides that, the reduced amount of samples present in the datasets did not appear to be an obstacle in generalization of the models for the investigated datasets. Due to its characteristics that favor the use of spectroscopy techniques in various everyday problems, a case study with data obtained from samples of patients with COVID-19 was carried out using 1D-CNN, PLS-DA and KNN for comparison. The algorithms were applied to a public dataset referring to samples of patients from the control group and patients infected with COVID-19. 1D-CNN together with the Savitzky–Golay filter obtained the best results and was used for comparison with other studies involving the diagnosis of COVID-19 from spectral data. The results presented show that approaches involving 1D-CNN's provided the best performance in terms of accuracy, specificity and sensitivity. In particular, in problems such as COVID-19 detection, a high sensitivity rate is very desired  and needed since false negatives represent the worst scenario. In this work, the  obtained results  were, on average, $96\%$ accuracy, $98\%$ specificity and $94\%$ sensitivity, indicating the feasibility of solutions for virus diagnosis using data obtained through spectroscopy with 1D-CNN as classifier. In future works, the influence of signal processing algorithms used in the data pre-processing step can be further investigated. Also, many problems present information beyond that the spectrum provides and may be utilized.

\section*{Acknowledgments}
R.A. Krohling thanks the Brazilian research agency Conselho Nacional de Desenvolvimento Científico e Tecnólogico (CNPq) - grant n.304688/2021-5.

\bibliographystyle{plainnat}
\bibliography{references}  %%% Uncomment this line and comment out the ``thebibliography'' section below to use the external .bib file (using bibtex) .

\end{document}